\documentclass[a4paper, 10pt, conference]{ieeeconf}      

\IEEEoverridecommandlockouts                              

\let\labelindent\relax
\usepackage[dvipdfmx]{graphicx}
\usepackage[frozencache,cachedir=minted-cache]{minted}
\usepackage{booktabs}
\usepackage{array}

\usepackage{enumitem}

\title{\LARGE \bf
Developing Interactive Tourism Planning: A Dialogue Robot System Powered by a Large Language Model*
}

\author{Katsumasa Yoshikawa$^{1}$ and Takato Yamazaki$^{1}$ and Masaya Ohagi$^{1}$ and Tomoya Mizumoto$^{1}$ and Keiya Sato$^{1}$
\thanks{$^{1}$SB Intuitions Corp.
        Tokyo Portcity Takeshiba Office Tower 1-7-1, Kaigan, Minato-Ku, Tokyo, 105-7529 Japan
        {\tt\small 
        katsumasa.yoshikawa@sbintuitions.co.jp}}%
}

\begin{document}

\maketitle
\thispagestyle{empty}
\pagestyle{empty}
\setlength{\abovedisplayskip}{1pt} 
\setlength{\belowdisplayskip}{1pt}

\setlength\floatsep{0pt}
\setlength\textfloatsep{0pt}
\setlength\intextsep{0pt}
\setlength\abovecaptionskip{0pt}
\begin{abstract}
In this paper, we propose a novel tourism planning system for Dialogue Robot Competition 2023 (a.k.a DRC2023).
\end{abstract}

\section{INTRODUCTION} \label{intro}
In recent years, large language models (LLMs) have rapidly proliferated and have been utilized in various tasks, including research in dialogue systems. 
We aimed to construct a system that not only leverages the flexible conversational abilities of LLMs but also their advanced planning capabilities to reduce the speaking load on human interlocutors and efficiently plan trips. 
Furthermore, we propose a method that divides the complex task of a travel agency into multiple subtasks, managing each as a separate phase to effectively accomplish the task.
Our proposed system confirmed a certain level of success by achieving fourth place in the DRC2023\cite{minato2023overview} preliminaries rounds. We report on the challenges identified through the competition.

\section{Proposed System}
Our system advances the conversation based on a pre-determined scenario-based approach.
Table \ref{tab:phases} shows the scenarios used, along with methods for controlling the robot's movements and the display.
The scenario consists of five \textit{phases} as shown in the table, each aimed at accomplishing a subtask in tourist guidance.
A large-scale language model is employed for response generation, and different types of prompts are used in each phase.
For the large-scale language model, the system uses the API of gpt-4-32k-0613 (referred to as GPT) provided by OpenAI.
We used the Streaming API and input the text into the Text-to-Speech (TTS) system every time a punctuation mark appeared, in order to speed up the speech delivery.
\newcommand{\thline}{\noalign{\hrule height 1pt}} 
\begin{table*}
    \centering
    \caption{Dialog phases of our system.}
    \label{tab:phases}
    \begin{tabular}{>{\raggedright}p{0.07\linewidth}>{\raggedright}p{0.15\linewidth}>{\raggedright}p{0.15\linewidth}>{\raggedright}p{0.15\linewidth}>{\raggedright}p{0.15\linewidth}>{\raggedright\arraybackslash}p{0.15\linewidth}} \toprule
         \textbf{Phase} & \textbf{1. Introduction \& Ice Breaker} &  \textbf{2. Inquiry} & \textbf{3. Course \& Spot Selection} & \textbf{4. Schedule Proposal} & \textbf{5. Confirmation \& Closing} \\ \midrule
         \textbf{Dialog Content} & Engage in casual conversation about travel. & Ask questions to determine preferences. & Decide on a model course and two spots through discussion. & Propose plan details and address customer queries. & Confirm the selected spots and exchange closing greetings. \\ \midrule

        \textbf{Physical Actions \& Displays} & 
        \begin{minipage}[t]{\linewidth}  
            \begin{itemize}[leftmargin=*,align=left]
            \raggedright
                \item Perform a bowing motion during the greeting
            \end{itemize}
        \end{minipage}
        &

        &
        \begin{minipage}[t]{\linewidth}
            \begin{itemize}[leftmargin=*,align=left]
            \raggedright
                \item Display images of the spots included in the model course
                \item Display an image of a spot
            \end{itemize}
        \end{minipage}
        &
        \begin{minipage}[t]{\linewidth}
            \begin{itemize}[leftmargin=*,align=left]
            \raggedright
                \item Display images and maps of the selected spots
            \end{itemize}
        \end{minipage}
        &
        \begin{minipage}[t]{\linewidth}
            \begin{itemize}[leftmargin=*,align=left]
            \raggedright
                \item Bowing motion at the end
            \end{itemize}
        \end{minipage}
        \\ \bottomrule
    \end{tabular}
\end{table*}

We implemented two types of transitions between phases. One is to instruct the system to output a termination sign [END] when the objective of the phase is achieved. When [END] is output, a transition is made to the next phase. This allows, for example, an automatic transition to the next phase when the system has heard everything it needs to hear in the \emph{inquiry} phase. However, the model often does not output [END], so the first transition condition may not be satisfied. To address this issue, we implemented a second transition that sets a maximum number of turns for each phase and forcibly terminates the phase if the number of turns is exceeded.

\subsection{Image and Map Viewer}
In this competition, monitors are set up to display information about tourist spots, including names, photos, and maps, provided through a designated API.
A web-based viewer is implemented for this display.
This viewer can show information up to 4 tourist spots.
Additionally, to aid those who might find it difficult to read the Kanji (Chinese characters) names of these tourist spots, phonetic readings (Furigana) are provided.

For displaying images, we implemented two types of timing. 
The first is to display the image when the name of a tourist spot appears in the system's utterance. If more than one spot appears in the same utterance, the images of all the spots are displayed. However, since more than four images cannot be displayed at the same time, only the fourth image on the display is replaced after the fifth image. The second timing is to display a list of images associated with the model course when the name of the model course is mentioned. Since this list of images is composed of images that symbolize the model course and are considered to be highly appealing to customers, it was displayed in preference to the first condition.

\subsection{Model Courses and Course Selection}
In the proposed system, model courses for a trip to Kyoto are collected from the internet in advance, and these are used to generate pairs of various personas and original model courses.
We then adopt a strategy to select and present the appropriate model courses from these candidates based on the content of the dialogue. 
This strategy reduces the decision-making burden on the customer's side and allows for the construction of a travel plan smoothly and in a short time.
Specifically, based on the dialogue history with the customer after \emph{inquiry} phase, our system selects the top two model courses suitable for the customer.

\begin{figure}[t]
    \centering
    \begin{minted}[breaklines, fontsize=\scriptsize, frame=single, breaksymbol=\quad, breaksymbolindentnchars=2]{bash}
Your name is Shoko...  # background and personas
Introduce two courses...  # task instruction
--- # Model courses information
Course A: The autumn leaves are... (Spot1-Spot2-...)  
Course B: Famous temples are...(Spot3-Spot4-...)  
=== # Shots
Shoko: ...
Customer: ...
... # Up to two shots
===  # Ongoing conversation
Shoko: ...  # Dialogue history
Customer: ...  
...
Shoko: # Start from here
    \end{minted}
    \caption{Main Prompt}
    \label{fig:main-prompt}
\end{figure}
\subsection{Model Course Introduction and Spot Selection}
After the model courses have been selected, the dialogue proceeds by introducing these two courses to the customer. This phase uses the main prompt, which is at the core of our system. This single prompt handles everything from indirectly hearing the customer's preferences through the dialogue introducing the model courses, to deciding on two tourist spots. The prompt extrapolates information from the two model courses determined in the previous phase, and instead of explaining all the information at once, it instructs to introduce tourist spots one by one within an interactive dialogue.

Furthermore, by using the Rurubu API, we continuously display images of tourist spots, receive customer reactions, and lead the dialogue up to the decision of course preference and the spots they liked within it. The end condition of this phase is the decision of two tourist spots. At the end of the phase, the dialogue history up to that point is input to the tourist spot extraction prompt, and the decided tourist spots are extracted in the backend.

\subsection{Route Finder and QA}
In order to create a schedule to visit two tourist spots, it is necessary to provide accurate route guidance that takes into account the latest traffic information. 
We achieve this by adding the latest route information obtained using the NAVITIME API\footnote{https://api-sdk.navitime.co.jp/api/} to the prompt.
API responses are converted into natural language using templates before being injected into the prompt.

The proposal of the schedule is also handled flexibly through generation, and the schedule is determined while confirming the customer's intentions. In addition to the route information, the prompt for this phase extrapolates basic information about tourist spots from the Rurubu API, so it can answer questions about required time, fees, and depending on the customer's request, information about lunch, among others.

\section{Evaluation}
For system evaluation, we used android robots set up at the storefronts of travel agencies in Fukuoka and Nagoya, and had people who visited the travel agencies evaluate our system.
The evaluation of the system was divided into two parts: the evaluation of the plan and the satisfaction evaluation.
Our system achieved a score of 0.85 in the plan evaluation, and an average of 4.41 across nine metrics in the satisfaction evaluation.
For the satisfaction evaluation, there was a large variance of 3.44, indicating a significant variation in perceptions among the participants.

\section{Related Work}

In the 2022 Dialogue Robot Competition (DRC2022) and before \cite{minato2023journal}, the task was to recommend one tourist destination from two options within a five-minute dialogue.
However, this year the competition has become more challenging due to the increased dialogue duration and a greater number of tourist destinations to choose from.
Previous works participated in DRC2022 \cite{yamazaki2022tourist, yamazaki2023building} implemented a scenario-based dialogue system and utilized an LLM for response generation.
This paper focuses on leveraging the performance improvements and enhanced controllability of large-scale language models, which have benefited from recent advancements like instruction tuning and scaling up.
To increase reliance on the LLM and introduce more flexibility in response generation, we have expanded the granularity of the scenarios.
Additionally, in response to the increased complexity of the task, we have also expanded our external knowledge sources, including the use of route searching APIs and model course datasets.

\section{Conclusion}
In this paper, we have outlined the proposed system that was entered in DRC2023 and discussed the issues discovered and insights gained through the competition.





\bibliographystyle{IEEEtran}
\bibliography{root}

\begin{thebibliography}{1}
\providecommand{\url}[1]{#1}
\csname url@rmstyle\endcsname
\providecommand{\newblock}{\relax}
\providecommand{\bibinfo}[2]{#2}
\providecommand\BIBentrySTDinterwordspacing{\spaceskip=0pt\relax}
\providecommand\BIBentryALTinterwordstretchfactor{4}
\providecommand\BIBentryALTinterwordspacing{\spaceskip=\fontdimen2\font plus
\BIBentryALTinterwordstretchfactor\fontdimen3\font minus \fontdimen4\font\relax}
\providecommand\BIBforeignlanguage[2]{{%
\expandafter\ifx\csname l@#1\endcsname\relax
\typeout{** WARNING: IEEEtran.bst: No hyphenation pattern has been}%
\typeout{** loaded for the language `#1'. Using the pattern for}%
\typeout{** the default language instead.}%
\else
\language=\csname l@#1\endcsname
\fi
#2}}

\bibitem{minato2023overview}
T.~Minato, K.~S. Ryuichiro~Higashinaka, T.~Funayama, H.~Nishizaki, and T.~Nagai, ``Overview of dialogue robot competition 2023,'' \emph{Proceedings of the Dialogue Robot Competition 2023}, 2023.

\bibitem{minato2023journal}
T.~Minato, R.~Higashinaka, K.~Sakai, H.~N. Tomo~Funayama, and T.~Nagai, ``Design of a competition specifically for spoken dialogue with a humanoid robot,'' \emph{Advanced Robotics}, vol.~37, no.~21, pp. 1349--1363, 2023.

\bibitem{yamazaki2022tourist}
T.~Yamazaki, K.~Yoshikawa, T.~Kawamoto, M.~Ohagi, T.~Mizumoto, S.~Ichimura, Y.~Kida, and T.~Sato, ``Tourist guidance robot based on hyperclova,'' \emph{arXiv preprint arXiv:2210.10400}, 2022.

\bibitem{yamazaki2023building}
\BIBentryALTinterwordspacing
T.~Yamazaki, K.~Yoshikawa, T.~Kawamoto, T.~Mizumoto, M.~Ohagi, and T.~Sato, ``Building a hospitable and reliable dialogue system for android robots: a scenario-based approach with large language models,'' \emph{Advanced Robotics}, vol.~37, no.~21, pp. 1364--1381, 2023. [Online]. Available: \url{https://doi.org/10.1080/01691864.2023.2244554}
\BIBentrySTDinterwordspacing

\end{thebibliography}

\end{document}